# Bridging Ethical Principles and Algorithmic Methods: An Alternative Approach for Assessing Trustworthiness in AI Systems

Michael Papademas[1,2], Xenia Ziouvelou[1], Antonis Troumpoukis[1], Vangelis Karkaletsis[1]

[1]National Centre for Scientific Research Demokritos,
Institute of Informatics and Telecommunications, Greece

[2]Panteion University of Social and Political Sciences,
Department of Communication, Media and Culture, Greece

{papademasmichael, xeniaziouvelou, antru, vangelis}@iit.demokritos.gr

## Abstract

Artificial Intelligence (AI) technology epitomizes the complex challenges posed by human-made artifacts, particularly those widely integrated into society and exerting significant influence, highlighting potential benefits and their negative consequences. While other technologies may also pose substantial risks, AI's pervasive reach makes its societal effects especially profound. The complexity of AI systems, coupled with their remarkable capabilities, can lead to a reliance on technologies that operate beyond direct human oversight or understanding. To mitigate the risks that arise, several theoretical tools and guidelines have been developed, alongside efforts to create technological tools aimed at safeguarding Trustworthy AI. The guidelines take a more holistic view of the issue but fail to provide techniques for quantifying trustworthiness. Conversely, while technological tools are better at achieving such quantification, they lack a holistic perspective, focusing instead on specific aspects of Trustworthy AI. This paper aims to introduce an assessment method that combines the ethical components of Trustworthy AI with the algorithmic processes of PageRank and TrustRank. The goal is to establish an assessment framework that minimizes the subjectivity inherent in the self-assessment techniques prevalent in the field by introducing algorithmic criteria. The application of our approach indicates that a holistic assessment of an AI system's trustworthiness can be achieved by providing quantitative insights while considering the theoretical content of relevant guidelines.

**Keywords:** AI Ethics, Trustworthy AI, Ethical Principles, AI System, PageRank, TrustRank, Philosophical Trust & Trustworthiness

## 1. Introduction

A great deal of effort has been put into creating systems that incorporate Artificial Intelligence (AI) methods in recent years. Algorithmic methods have been used for several years, but significant improvements in computational power and extensive application of data collection techniques have led to the development of easy-to-use and efficient AI systems. To ensure the efficiency of the systems, quite complex processes are run in the background, using large amounts of data, advanced algorithms, and high computational power [1]. Although these modern applications make it easier for individuals and organizations to carry out their daily tasks, sometimes the results of AI systems are confusing, unclear, even though plausible and realistic. Because of the algorithmic and computational complexity, the individual cannot fully understand the reasoning of the system and how it makes decisions and predictions.

The difficulty people face in effectively and ethically integrating these systems into their lives can lead to inconsistent use of AI, creating high-risk situations from both social and individual perspectives. The distorted implications of AI systems have as their main source the ignorance of people about issues related to the ethical use of technologies. A well-known example of deviation from ethical and trustworthy use is Amazon's employee recruitment software. By 2015, the company realized that the recruitment system was not rating candidates for technical jobs and posts in a gender-neutral way [2]. Also, the recidivism algorithm used in U.S. courts to predict the probability of re-offending was biased against black people [3]. In the previous two examples, we saw how an untrustworthy algorithmic system can affect one's professional career or even the legal right to equal treatment vis-à-vis the judicial system. In the following sections, we present the aspects and principles of AI Trustworthiness in our attempt to deconstruct a concept that seems ambiguous and chaotic.

As AI systems become more complex and opaque, our ability to fully comprehend and guide their behavior diminishes, increasing the risk of applications that unintentionally conflict with broad human principles, such as fairness, transparency, and accountability. An extension of this situation was the shift of political institutions towards understanding and institutionalizing regulatory rules. In 2018, the European Union (EU) proposed ethical guidelines for Trustworthy AI to govern and facilitate the development and operation of AI systems [4]. In 2021, the U.S. Government Accountability Office (GAO) published a framework for the accountability and responsible use of AI, identifying key practices to help ensure these aspects of AI [5]. Beyond institutional frameworks, research activity in the area of Trustworthy AI has been steadily increasing.

In recent years, organizations and academic communities have made a plethora of efforts to develop frameworks and methods for assessing the trustworthiness of AI systems, aiming to ensure these technologies are reliable, transparent, and aligned with ethical standards. Looking at the relevant literature, we could say that there is a saturation of institutional frameworks on Trustworthy AI. One of the major problems in the field of Trustworthy AI is the chasm that exists between theoretical frameworks and practical methods. The disproportionate emphasis on the theoretical frameworks and the neglect of practical methods may mean that some of the respective communities approach the issue of Trustworthy AI from an ethicswashing perspective. Ethicswashing refers to a strategic practice in which an organization presents a misleading and superficial commitment to ethical standards without implementing any meaningful or substantive actions [6]. Therefore, the need to bridge theory with practice is critical and necessary if we want to resolve the problem at its core.

In this paper, we approach the notion of trustworthiness assessment of AI systems from a different perspective. Aiming to overcome the subjective nature of AI self-assessment processes, we propose an alternative approach that appears to be a less subjective way of assessing AI trustworthiness. This approach employs existing algorithmic methods that inherently include an exploratory dimension, allowing insights into the AI system's components to be derived by examining their interrelationships and dependencies. More specifically, we explore the use of Link Analysis algorithms, such as PageRank and TrustRank, to elicit the trustworthiness level of various Trustworthy AI requirements, their key aspects, and the components of each aspect for AI systems under review. Our approach is designed not only to assess the trustworthiness of an AI system but also to offer insights into the mechanisms by which trust propagates across its various components, an area that, to the best of our knowledge, remains relatively underexplored in existing research. Thus, we utilize well-established, non-algorithmic Trustworthy AI frameworks, such as ALTAI [7], and integrate Link Analysis algorithms to explore a novel approach for assessing the trustworthiness of AI systems.

Toward this aim, the paper begins by exploring the notion of Trustworthy AI and presents a meta-analysis of systematic and semi-systematic literature reviews, intending to identify the basic ethical requirements that a Trustworthy AI system should fulfill, as discussed in Section 3. This theoretical foundation informs our pragmatic approach, which focuses on implementing two Link Analysis algorithms: PageRank and TrustRank. Section 4 provides an overview of existing Trustworthy AI assessment methods and tools from both theoretical and technological perspectives, all grounded in the concept of trust. In Section 5, we propose an alternative algorithmic approach for assessing the trustworthiness of AI systems, utilizing the aforementioned Link Analysis techniques to complement and enhance existing assessment frameworks. Additionally, we present an exploratory analysis that initially validates this proposed approach. In Section 6, we conclude the paper by discussing key findings, limitations, and potential directions for future research.

## 2. Methodology

To explore the ethical principles underlying Trustworthy AI, we conducted a scoping review of the existing literature, with the aim of identifying systematic and semi-systematic reviews that addressed these principles in a broad, cross-disciplinary context. Our primary objective was to synthesize ethical principles that transcend specific application domains, providing a comprehensive foundation for further research.

We began by conducting a comprehensive review of the available literature on AI ethics, focusing on studies that addressed ethical principles at a general level. However, we observed that many results focused on domain-specific applications, such as education and healthcare, rather than addressing ethical principles at a general level. To ensure the scope remained aligned with our objectives, we prioritized studies with a cross-disciplinary focus, excluding those heavily tied to specific domains. Building upon this foundation, we employed a targeted search strategy to complement our initial findings.

During this process, we noted a significant alignment between our findings and those presented in the meta-review by Ziouvelou, Giouvanopoulou, and Karkaletsis [8], likely due not only to the relatively short time frame between the two studies but also to the overlap in key research questions addressed in both bodies of work. Their work served as a valuable reference point, prompting a re-examination and extension of their dataset. This involved applying updated inclusion and exclusion criteria to ensure the relevance and currency of our analysis. Specifically, we excluded one empirical study from their original dataset, as it did not meet the criteria for systematic or semi-systematic reviews. In addition, we incorporated a new review published in 2024, which expanded the temporal coverage of the analysis. These steps ensured our review incorporated the most up-to-date and relevant literature.

The scoping review provided a body of papers, which we used to conduct a meta-analysis that offered a framework for identifying and synthesizing ethical principles across diverse contexts. The insights derived from this review established the conceptual foundations upon which our algorithmic approach was developed, ensuring that it aligns with the requirements of Trustworthy AI. In essence, the methodological approach adopted enables the integration of algorithmic methods, PageRank, and TrustRank within this theoretical framework.

# 3. Ethical Principles in Trustworthy AI

## 3.1 Literature Review and Meta-analysis

The ethics of AI and its social implications have attracted serious attention, and many policy frameworks and guidelines have been developed by various organizations [9, 10]. Additionally, each framework contains different AI ethics principles, which reflect unique perspectives on current and future AI strategies. The AI ethical guidelines serve to indicate to legislators that internal self-governance in science and industry is sufficient and that specific laws are not needed to mitigate potential technological risks and eliminate abuse scenarios [11]. It must not be neglected that ethical principles for developing and using AI systems are proposed and contextualized in countless AI ethics guidelines, which prescribe ethical direction to AI systems' developers, users, policymakers, and other stakeholders who seek to maximize the potential benefits, minimizing at the same time the potential harms of systems' operations [12]. Therefore, we consider it important to see the convergence of these principles in order to establish a set of minimum requirements regarding the trustworthiness of AI systems. We approach the aforementioned issue by comparing various systematic or semi-systematic reviews that have been conducted in the past on the ethical principles that should govern AI systems.

We conducted a meta-analysis of selected systematic literature reviews (SLRs) to synthesize existing insights into the ethical principles underlying Trustworthy AI. One of the first SLRs was that of Floridi et al. (2018) [13], which is actually a semi-SLR. They assessed six documents with a chronological range from 2017 to 2018, which yielded 47 principles and derived five high-level principles (Beneficence, Non-Maleficence, Autonomy, Justice, Explicability). The principle of Explicability was added by authors as a need to incorporate both intelligibility and accountability. Zeng, Lu, and Huangfu (2018) [9] collected 27 proposals of AI principles from societal entities like Academia, Non-profits and Non-Governmental Organizations, Governments, and Industry. Thinking about the issue from a semantic perspective, they posed a set of manually chosen core terms keywords, setting the core terms of Accountability, Privacy, Fairness, Humanity, Collaboration, Share, Transparency, Security, Safety, AGI/ASI (Artificial General / Super Intelligence). Jobin, Ienca, and Vayena (2019) [14] conducted a systematic scoping review of the existing corpus of documents containing soft-law or non-legal norms issued by organizations. In particular, they identified 84 eligible, non-duplicate documents containing ethical principles for AI, revealing a global convergence emerging around five ethical principles (Transparency, Justice and Fairness, Non-maleficence, Responsibility, and Privacy).

In the current decade, Fjeld et al. (2020) [15] analyzed the content of 36 prominent AI principles documents to identify trends and essential components in the discussion on the future of AI technologies. They established 47 principles governing the AI, which can be categorized into eight themes: Accountability, Privacy, Fairness & Non-discrimination, Safety & Security, Transparency & Explainability, Human Control of Technology, Professional Responsibility, Promotion of Human Values. They also point out that the aforementioned themes serve as fundamental requirements, expressing the importance of these conceptual and ethical principles. Hagendorff (2020) [11] conducted a semi-systematic literature review that compares 22 guidelines, finding that principles of Accountability, Fairness, and Privacy appear altogether in about 80% of all guidelines, and identifying 22 ethical principles in total. He also notes that these three core aspects constitute of minimum requirements for building and using AI systems ethically. Franzke (2022) [10] conducted research on a total of 70 AI ethics guidelines. The analysis concluded that the most dominant principles are Transparency, Privacy, and Accountability. She notes that AI ethics guidelines largely ignore the crucial question of how ethical principles can be transposed onto the usage of technology. Furthermore, Khan et al. (2022) [16] presented a systematic literature review revealing a global convergence around 22 ethical principles. Through their approach, Transparency, Privacy, Accountability, and Fairness are identified as the most common AI ethics principles. They highlight the existence of significant practical challenges involved in implementing the guidelines in real-world conditions, mentioning the lack of tools or frameworks that bridge the gap between principles and practice. Attard-Frost, Ríos, and Walters (2023) [12] concentrate on four a priori principles, which they have designated F.A.S.T., examining 47 AI ethics guidelines. These principles are Fairness, Accountability, Sustainability, and Transparency. Corrêa et al. (2023) [17] took into account 200 AI guidelines, identifying 17 principles prevalent in the respective policies and guidelines. From those principles, the top five, based on citation index, were found to be similar to the ones identified by Jobin et al. and Fjeld et al. Laine, Minkkinen, and Mäntymäki (2024) [18] conducted a systematic literature review to understand ethical principles and stakeholders in ethics-based AI auditing. From the sample of 110 studies, they conducted backward citation chaining. Finally, they assessed 93 articles on ethics-based AI auditing. The results were the grouping of 54 terms related to ethics into eight principles: Fairness, Transparency, Non-Maleficence, Responsibility, Privacy, Trust,

Beneficence, Freedom and Autonomy. The authors enriched the results of the systematic literature review by adding three additional principles, namely Sustainability, Dignity, and Solidarity, in order to enhance comprehensiveness, even though none of the studies explicitly mentioned them. Table 1 presents the results of the systematic literature reviews, while Figure 1 presents the outcomes of the meta-analysis.

**Table 1:** Dominant Ethical Principles in Meta-Analyzed Systematic Reviews.

| Year | Author(s) | Study Type | No. of Guidelines / Studies | Total Ethical Principles | Converging Ethical Principles | Converging Key Ethical Principles / Themes |
|---|---|---|---|---|---|---|
| 2018 | Floridi et al. | Semi-SLR | 6 | 47 | 5 | Beneficence, Non-Maleficence, Autonomy, Justice, Explicability |
| 2018 | Zeng et al. | Semi-SLR | 27 | - | 10 | Accountability, Privacy, Fairness, Humanity, Collaboration, Share, Transparency, Security, Safety, AGI/ASI |
| 2019 | Jobin et al. | S(Sc)LR | 84 | 11 | 5 | Transparency, Fairness, Non-Maleficence, Responsibility, Privacy |
| 2020 | Fjeld et al. | SLR | 36 | 47 | 8 | Accountability, Privacy, Fairness & Nondiscrimination, Safety & Security, Transparency & Explainability, Human Control of Technology, Professional Responsibility, Promotion of Human Values |
| 2020 | Hagendorff T. | Semi-SLR | 22 | 22 | 3 | Accountability, Privacy, Fairness |
| 2022 | Khan et al. | SLR | 27 | 22 | 4 | Transparency, Privacy, Accountability, Fairness |
| 2022 | Franzke A. | SLR | 70 | - | 4 | Transparency, Privacy, Accountability, Safety |
| 2023 | Attard-Frost et al. | Semi-SLR | 47 | 4 | 4 | A priori classification based on the FAST (Fairness, Accountability, Sustainability, Transparency) |
| 2023 | Correa et al. | SLR | 200 | 17 | 6 | Transparency/Explainability/Auditability, Reliability/Safety/Security/Trustworthiness, Justice/Equity/Fairness/Non-discrimination, Privacy, Accountability/Liability, Freedom/Autonomy/Democratic Values/Technological Sovereignty |
| 2024 | Laine et al. | SLR | 93 | 54 | 8 | Fairness, Transparency, Non-Maleficence, Responsibility, Privacy, Trust, Beneficence, Freedom and Autonomy |

SLR: Systematic Literature Review, S(Sc)LR: Systematic Scoping Literature Review

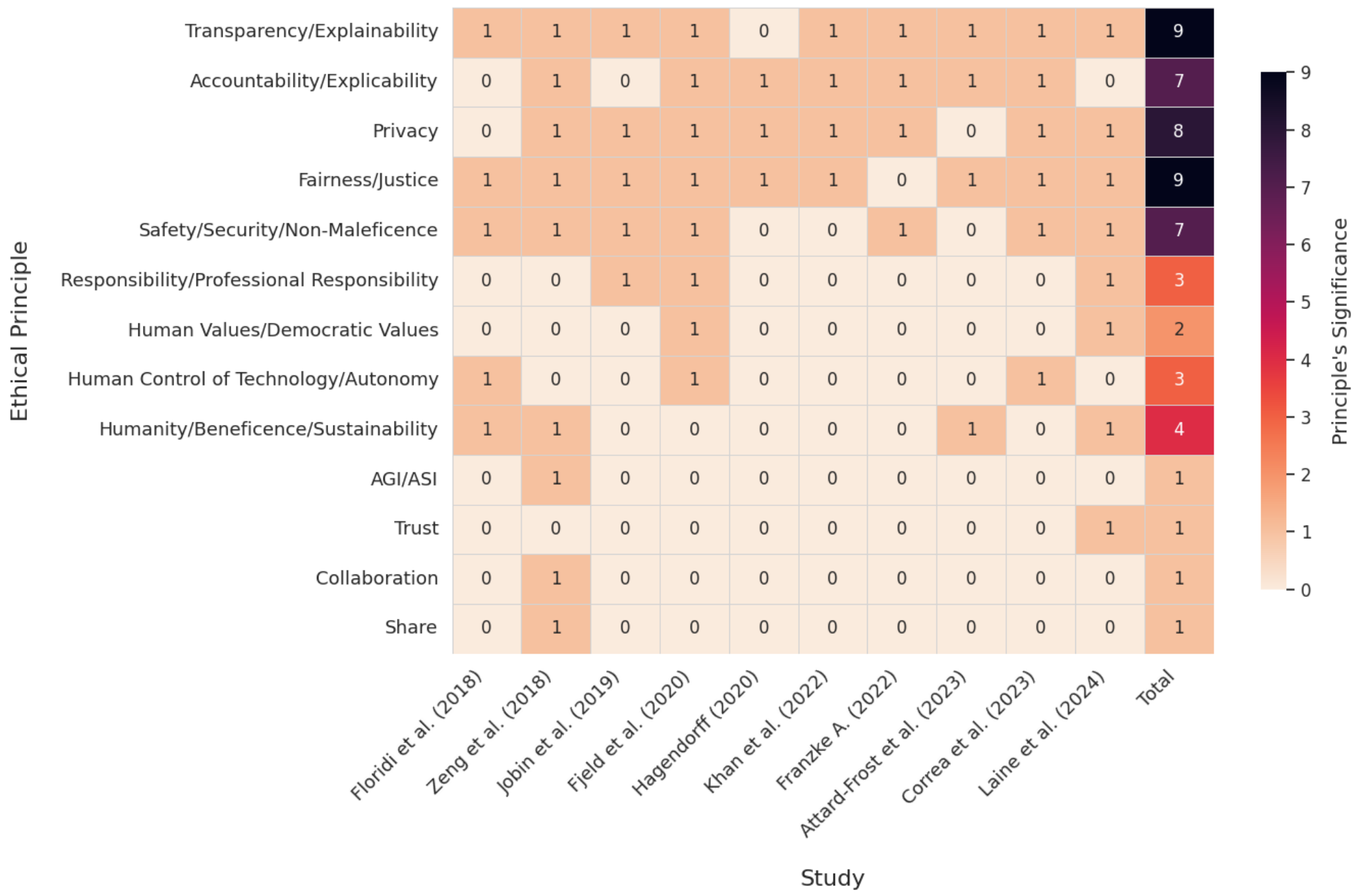

**Figure 1:** Meta-analysis Results as a Synoptic Heatmap of AI Ethics Principles.

## 3.2 Convergence of Meta-analysis Results with AI HLEG

It is noteworthy that the principles of Trustworthy AI, as delineated in the meta-analysis, appear to be consistent with the corresponding requirements set forth by the High-Level Expert Group on AI (AI HLEG), which was among the first to address the issue of Trustworthy AI in a formal document. The AI HLEG Trustworthy AI Guidelines have been formulated as non-legal and non-binding guidelines to direct the development of AI, taking into account a wide range of ethical principles in an effort to balance innovation and safety [19]. The European Commission (EC) communicated an AI strategy in 2018. In 2019, the AI HLEG published the Ethics Guidelines for Trustworthy AI, and in 2020 published the Assessment List for Trustworthy AI (ALTAI) [7, 20], which translates the seven key requirements into several checklists. The document presented by the AI HLEG lists seven requirements that must be met for AI systems to be trustworthy.

In particular, the EU High-Level Expert Group's guidelines define that Trustworthy AI has three dimensions: Lawfulness, Ethicalness, and Robustness. The AI system should be Lawful, being in alignment with all applicable laws and regulations. Furthermore, it needs to be Ethical, ensuring that ethical principles and values are upheld. Finally, it should be Robust both from a technical and social perspective. The ethical dimension of the system seems especially important, specifying four ethical principles:

Respect for Human Autonomy, Prevention of Harm, Fairness, and Explicability. The Trustworthy AI is realized through the seven requirements: Human Agency & Oversight, Technical Robustness & Safety, Privacy & Data Governance, Transparency, Diversity, Non-Discrimination & Fairness, Societal & Environmental Wellbeing, and Accountability.

In the meta-analysis review, we identify the principles on which there is convergence across the systematic and semi-systematic literature reviews. We select the seven principles with the highest frequency, as shown in Figure 1, to assess their alignment with the seven AI HLEG requirements. Table 2 shows the overlap between the results of the meta-analysis and the requirements outlined in the EU guidelines. It is observed that the ALTAI requirements are encapsulated within the seven most essential principles derived from the meta-analysis. The requirement of "Human Agency & Oversight" aligns with the principle of "Human Control of Technology/Autonomy", while "Technical Robustness & Safety" corresponds to "Safety/Security/Non-Maleficence". Similarly, the need for "Data Privacy & Governance" is addressed by the principle of "Privacy", and the requirement of "Transparency" aligns with "Transparency/Explainability". Moreover, "Diversity, Non-Discrimination & Fairness" is conceptually linked to "Fairness/Justice", and "Societal & Environmental Wellbeing" connects with "Humanity/Beneficence/Sustainability". Finally, the principle of "Accountability/Explicability" encompasses the requirement of "Accountability". Although "Responsibility/Professional Responsibility" is one of the seven highest-scoring principles identified in the meta-analysis, it is not among the AI HLEG requirements. In the context of this paper, we rely on the seven requirements set forth by the High-Level Expert Group, as there is a clear alignment with the findings of our meta-analysis.

These theoretical requirements serve as the foundation for constructing the conceptual framework underlying our graph-based approach, where each requirement informs a distinct node or set of nodes within the graph. This structure enables the application of Link Analysis algorithms to model and quantify trust relationships within the conceptual world of an AI system.

**Table 2:** Alignment Between the Results of the Meta-analysis and the Requirements of the EU about Trustworthy AI.

| **Requirement for Trustworthy AI (ALTAI - AI HLEG)** | **Description of Requirement [1, 7, 20]** | **Appearance in the meta-analysis** |
|---|---|---|
| Human Agency & Oversight | This necessitates that AI systems should serve as facilitators for a democratic, thriving, and just society, enabling user autonomy and upholding fundamental rights, while also allowing for human supervision. | **Human Control of Technology / Autonomy** |
| Technical Robustness & Safety | This is closely linked to the principle of harm prevention. The development of AI systems must be undertaken with a preventative approach to risks, and in a manner that ensures the reliable behavior of the system in question while minimizing the potential for unintentional and unexpected harm, and preventing any instances of unacceptable harm. | **Safety/Security / Non-Maleficence** |
| Privacy & Data Governance | The prevention of harm to privacy is contingent upon the implementation of robust data governance frameworks that encompass the quality and integrity of the data utilized, its relevance in the context of the domain in which the AI systems will be deployed, its access protocols, and the capacity to process data in a manner that safeguards privacy. | **Privacy** |
| Transparency | This requirement is closely linked with the principle of explainability and encompasses the transparency of elements relevant to an AI system, including the data, the system itself, and the business models. | **Transparency / Explainability** |
| Diversity, Non-Discrimination & Fairness | AI systems should treat all sections of society fairly without discriminating based on factors such as socio-economic determinants. They should not cause any direct or indirect discrimination to any group of society. This requirement enables the AI system to be available and accessible to all sections of society without any discrimination. | **Fairness / Justice** |
| Societal & Environmental Wellbeing | AI systems should not cause any harm to society or the environment during their design, development, and use. Overall, AI should be used to benefit all human beings, including future generations. AI systems should serve to maintain and foster democratic processes and respect the plurality of values and life choices of individuals. | **Humanity / Beneficence / Sustainability** |
| Accountability | The principle of accountability necessitates that mechanisms be established to ensure responsibility for the development, deployment, and/or use of AI systems. This topic is closely related to risk management, which involves identifying and mitigating risks in a transparent manner that can be explained to and audited by third parties. | **Accountability / Explicability** |

# 4. Trustworthy AI Assessment Methods and Tools

## 4.1 Theoretical Tools and Frameworks

The importance of Trustworthy AI assessment in the context of development and deployment is increasingly recognized, with a variety of methodologies and tools being utilized to evaluate and ensure adherence to trustworthiness principles. In this context, one of the earliest and most widely recognized frameworks is the ALTAI [7, 20]. Developed by the AI HLEG, ALTAI serves as a self-assessment tool that provides a structured and practical approach for evaluating AI systems in alignment with ethical guidelines. The framework provides guidance on the fundamental pillars of Trustworthy AI. By outlining seven key requirements, it underscores the essential aspects that organizations must address to ensure their AI systems are trustworthy.

The Organisation for Economic Co-operation and Development (OECD) established its AI Principles in 2019 [21], which provided a foundational international agreement on fostering AI systems that are innovative, sustainable, and beneficial to society. To establish a foundation for global interoperability between jurisdictions, countries adopt the OECD AI Principles and associated tools to formulate policies and establish AI risk frameworks.

The World Economic Forum (WEF) has also contributed with its framework, released in 2020 [22], which emphasizes embedding ethical considerations throughout the AI lifecycle to build public trust and ensure global impact. The Forum's trust framework demonstrates how key principles such as cybersecurity, privacy, transparency, redressability, auditability, fairness, interoperability, and safety can enhance trust in technology and the organizations that develop and use it. The accompanying report offers a structured framework and actionable roadmap for fostering trustworthiness in the development and application of technological systems.

Likewise, the MITRE developed the AI Maturity Model and Organizational Assessment Tool Guide in 2023 [23], providing organizations with a structured approach to evaluate their AI systems' maturity across dimensions such as governance, risk management, and operational effectiveness. This assessment tool is designed to operationalize the maturity model, offering organizations valuable insights and a clear understanding of the critical areas required to support the development and advancement of AI technologies.

Additionally, the National Institute of Standards and Technology (NIST) published its AI Risk Management Framework in 2023 [24], with a particular emphasis on risk mitigation and the promotion of reliable AI outcomes through robust governance practices. Explains the purpose of the AI Risk Management Framework (RMF), which is to provide organizations with guidance on managing AI risks and promoting Trustworthy AI development and use, describing the potential benefits and risks of AI technologies.

## 4.2 Technological Tools

At the same time, a variety of specialized tools have been developed to quantify and address key dimensions of AI trustworthiness. For fairness, tools such as AIF360 [25] and scikit-lego provide robust mechanisms to identify, quantify, and mitigate biases in datasets and machine learning models. These tools play a crucial role in fostering equitable AI systems by addressing disparities that may arise from skewed or incomplete data.

Related to the requirement of robustness, frameworks such as the Adversarial Robustness Toolbox (ART) [26] and secml [27] offer advanced capabilities to evaluate AI systems' resistance to adversarial attacks. These tools are essential for ensuring that AI systems perform reliably even under malicious or unforeseen perturbations. By simulating adversarial scenarios, these frameworks help improve system resilience and build user confidence in real-world applications.

Considering explainability, tools like AIX360 [28] and secml are instrumental in improving the interpretability of AI models. They enable stakeholders to understand how models arrive at their predictions, implicitly ensuring aspects related to requirements such as transparency and accountability. Also, they include algorithms that cover the different dimensions of explanation modes along with proxy explanation metrics.

Another vital aspect of AI trustworthiness is uncertainty quantification, addressed by tools like UQ360 [30]. This tool provides methods to measure and manage the confidence of AI predictions, enabling decision-makers to appropriately weigh the reliability of model outputs. Furthermore, offers a comprehensive suite of tools to streamline and enhance the practices of quantifying, evaluating, improving, and communicating uncertainty throughout the AI application development lifecycle [30]. Additionally, it promotes deeper exploration of uncertainty's connections to other aspects of Trustworthy AI, such as fairness and transparency, by sharing cutting-edge research and educational resources.

It is important to acknowledge the ongoing efforts to develop tools that integrate assessment methods for multiple AI trustworthiness requirements [30]. The commitment to pursuing more holistic solutions represents a highly optimistic aspect in addressing the challenge of Trustworthy AI and the strategies for its realization.

## 4.3 The Nature of Existing Assessment Methods

Based on the tools reviewed so far, it is clear that methods for assessing Trustworthy AI generally fall into two broad categories. The first category includes tools that are grounded in ethical principles and provide guidelines on how to incorporate the positive attributes of relevant technologies while mitigating their potential negative impacts throughout the life cycle of the AI system. These tools take a holistic perspective, addressing both the technological and social dimensions of AI systems, although they often lack mechanisms for precise quantification. The second category includes programming frameworks and technological toolkits that use algorithms and metrics to identify and assess specific characteristics of AI systems. While these practical tools excel at quantifying certain aspects of Trustworthy AI, they tend to focus on narrower dimensions of trustworthiness. While both categories incorporate self-assessment methods to varying degrees, such tools and methods inherently introduce an element of subjectivity into the evaluation process. The approach we propose mitigates the subjective nature of self-assessment by incorporating algorithmic techniques while ensuring that their operation remains grounded in qualitative features defined by the ethical requirements of the AI HLEG.

# 5. Assessing AI System's Trustworthiness through PageRank and TrustRank

## 5.1 Trust and Trustworthiness

The concepts of trustworthiness and trust are multifaceted, and they can be understood in various ways, depending on one's perspective. In the context of AI, it is the idea of a framework ensuring a system is trustworthy, based on evidence of its stated requirements. It ensures that the expectations of users and stakeholders are met in a verifiable way [1, 31]. The conceptual basis of trustworthiness lies in the concept of trust. Kaur et al. [1] note that sociologists see trust as relational, psychologists as cognitive, and economists as calculative. Given the subtlety of this distinction and the fact that it can sometimes be overlooked in more technical or functionally driven approaches, it is essential to briefly examine the various dimensions of trust and trustworthiness.

Returning to the issue of definitions, economists tend to reduce trust to calculative expectations or institutional guarantees [32], framing it as a form of rational risk-taking. Psychologists, meanwhile, focus on internal cognitive models of the trustor and perceived attributes of the trustee [33]. Sociologists may treat trust as a function of social embedding or institutional structures [33]. In the management sciences, trust is defined behaviorally as "the willingness of a party to be vulnerable" in relationships involving risk and uncertainty [34]. In the context of informatics and artificial intelligence, trust is often understood as the trustor's willingness to rely on a system's ability to perform specific actions or provide services within a defined context [35]. This reliance is shaped by beliefs about the system's competence, integrity, benevolence, and predictability [36], and reflects a readiness to accept vulnerability in technologically mediated interactions. Despite their disciplinary differences, these definitions converge on the idea that trust entails accepting vulnerability, supported by perceptions of competence, integrity, and reliability.

While these definitions are contextually appropriate, they often rely, implicitly or explicitly, on a philosophical core. Both phenomena are associated with a certain degree of expectation, vulnerability, and moral hope. These concepts are examined in greater detail in philosophical discourse. The Oxford Dictionary of Philosophy defines trust as "the attitude of expecting good performance from another party" [37], tying it closely to values like loyalty, truthfulness, and promise-keeping. Philosophically, trust can be understood as the confidence one entity has in another that the latter will behave as expected. It is necessary to point out that trust is an attitude that we have towards entities in which there is the hope that they will be trustworthy, where trustworthiness is a property and not an attitude [38]. Trust involves a form of reliance, but it goes beyond simple dependence, including some extra factor [39, 40, 41]. This extra factor generally concerns why the trustor (i.e., the one who trusts) would rely on the trustee (i.e., the one who is trusted) to be willing to carry out the actions for which they have been trusted.

The philosophical definition is especially valuable because it highlights the normative dimension of trust. Trust is not simply a pragmatic strategy; it is a moral commitment. To trust someone or something is not only to rely on them but also to believe that they ought to act in a trustworthy manner. This normative link is what makes trust fragile, and its violation morally significant. While definitions of trust vary across disciplines, the philosophical approach offers a rich, unified, and normative framework that is either embedded within or provides the scaffolding for those other understandings. Far from being merely abstract, the philosophical conception of trust is indispensable for coherent interdisciplinary work.

Crucially, this moral and normative dimension also underpins trust in artificial intelligence and informatics. Trust in AI is not merely about functional performance or reliability; it implicitly assumes that systems will behave in ways aligned with human expectations and ethical standards. The philosophical account of trust thus provides the conceptual foundation to understand trust in AI not only as rational reliance but also as a form of moral engagement. In this context, reliance refers to the trustor's dependence on a system's ability to consistently fulfill its expected functions within a specific setting. Importantly, such reliance can occur not only between a human trustor and a technological system, but also among components within the system itself, where one component or ethical requirement depends on another to operate in a trustworthy manner and align with broader system goals. However, unlike traditional software, many AI systems, particularly those based on machine learning, derive their algorithms directly from data rather than explicit human instructions. This reliance on data-driven learning underscores the need for safeguards to ensure that these models are fit for purpose, ethically sound, and free from unintended biases, as their performance is fundamentally shaped by the quality and representativeness of the training data. Recognizing this, we propose an algorithmic approach to quantify these reliance relationships, aiming to interpret this extra factor within the qualitative context of established ethical frameworks, such as the ALTAI, discussed in the literature.

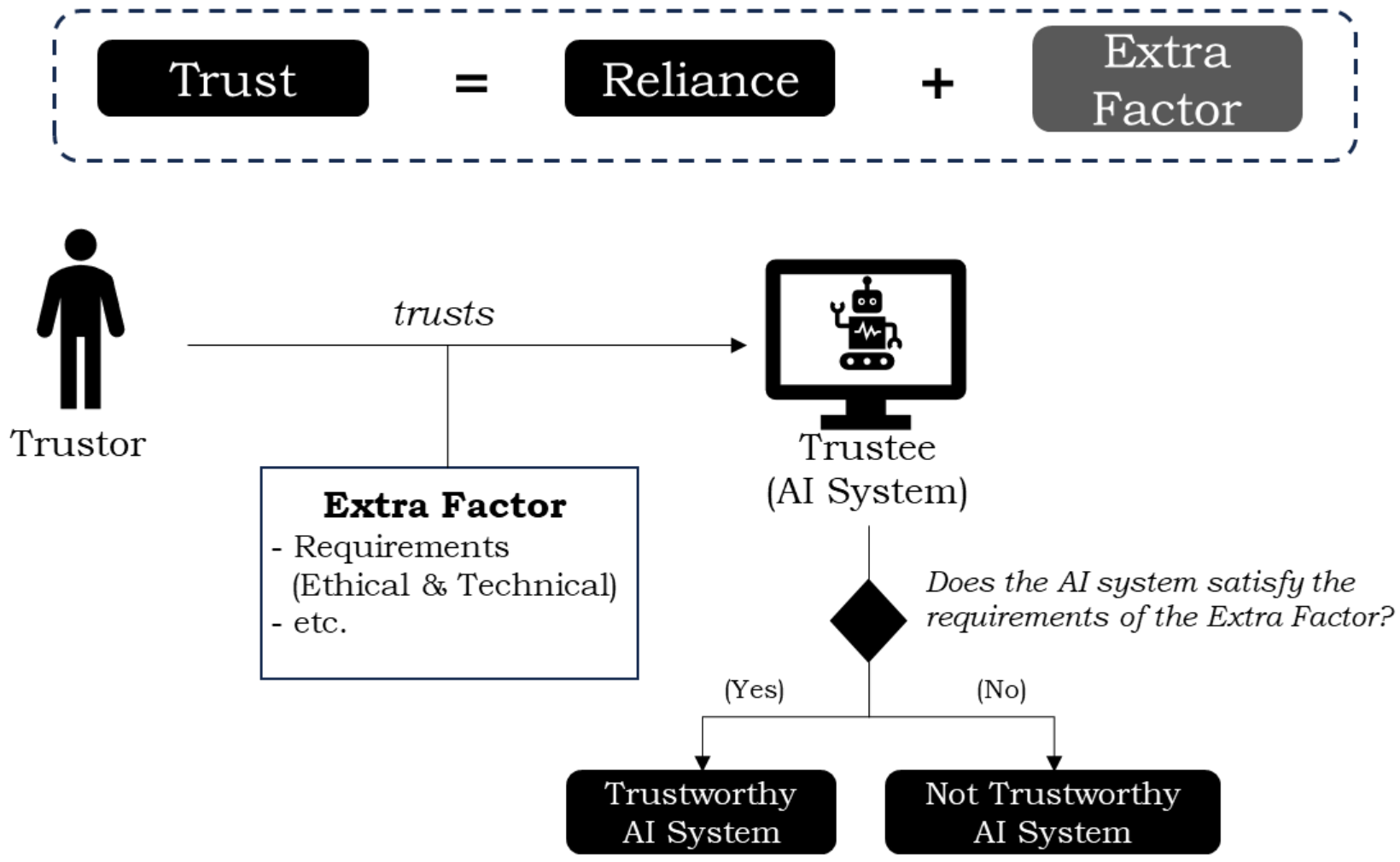

**Figure 2:** Trust and Trustworthiness in the AI Context.

## 5.2 PageRank and TrustRank

The above definitions contribute to a theoretical understanding of the concept, but they lack clear and actionable guidance for the practical implementation of trustworthiness, particularly in the context of AI. In 1999, Lawrence Page, describing the PageRank algorithm, which is a method for objectively and mechanically rating websites, noted that PageRank could help a user decide if a site is trustworthy or not [42]. These words give us the impetus to take a deeper look at the present algorithm and its extension, called TrustRank, proposing both algorithms as a solution to bridging the gap between theoretical and practical approaches to Trustworthy AI.

PageRank's main focus is to measure the importance of a webpage through the analysis of its inbound links. The core idea is that a page is important if many other important pages link to it [43, 44]. PageRank plays a central role in search engines as it reflects not only the popularity of a webpage but also serves as a rough approximation of its reliability and quality. This is based on the principle that pages linked to by many others are generally regarded as reliable and valuable sources of information, as the linking pages themselves likely possess some level of authority and credibility. However, while this approach captures a form of collective trust, it may not always fully account for nuanced dimensions of trustworthiness, such as the accuracy, intent, or ethical alignment of the content. PageRank's thesis is that a webpage is important if it is pointed to by other important pages [45, 46]. Furthermore, PageRank does not define or consider trust explicitly. It assumes that links themselves indicate a form of confidence from one page to another, with no distinction between trustworthy or untrustworthy links, but taking into account the popularity importance. The algorithm computes the probability that a random web surfer lands on a particular page by following links.

Pages with more inbound links from important or highly linked pages are given a higher PageRank score [43]. Also, it requires a complete analysis of the link structure of the web or graph. This involves indexing the web to understand how pages are linked to each other. Of course, the algorithm is not without its drawbacks, the most important being that it fails to differentiate between trustworthy and spammy pages. Spam can rank highly if it receives enough inbound links, since the algorithm's approach does not incorporate any knowledge about the quality of a site [47], nor does it explicitly penalize badness.

The challenge of defining trustworthiness in practical contexts brings the TrustRank algorithm to the forefront as a relevant methodological approach. In 2004, Gyöngyi, Garcia-Molina, and Pedersen developed the TrustRank, shifting the focus towards trustiness and potentially to trustworthiness. The intuition behind TrustRank is that a page with high PageRank, but without a relationship with any of the trusted pages, is suspicious [48] and, by extension, less trustworthy. The algorithmic approach begins by assuming that trusted sites, chosen by humans, are unlikely to link to untrusted or spammy sites. So, the selection of seed nodes by human structures the definition of trust. TrustRank begins with a set of manually selected seed pages that are known to be trustworthy. It then propagates trust scores from these seed pages to other linked pages, setting that pages closer to the seed pages receive higher trust scores. The limitations of this approach are related to the subjectivity introduced into the model by the human factor. Just as with email spam, determining if a page or group of pages is spam is subjective. However, we remain skeptical that a fully autonomous mechanism for evaluating the trustworthiness of another system can exist independently of human oversight, given the inherently anthropocentric nature of trust and the epistemic challenges involved.

After describing the algorithms, it is equally essential to consider the algorithmic process of these methods. In Table 3 we describe the stages of PageRank, with inputs being the set of pages (P), the damping factor ($\alpha$), and the convergence threshold ($\epsilon$). The damping factor prevents the algorithm from getting stuck in infinite loops. The intuition behind this factor is that a higher value means more weight to the link structure, while a lower value gives more weight to random jumps. Also, the factor is typically set to 0.85 [43], meaning there is an 85% chance the user will continue clicking on links and a 15% chance they will stop and start over from a random page, ensuring that the PageRank values converge. The convergence threshold determines the level of accuracy required, with a smaller threshold meaning more precise calculations, but longer computational time.

TrustRank, as an extension of PageRank, generally has a similar structure. Noteworthy is that the intuition behind the decay factor ($\alpha$) indicates that a higher value of the factor gives more weight to the link structure, while a lower value gives more weight to the initial trust assigned to seed pages. Furthermore, Table 4 illustrates the phases of TrustRank.

**Table 3:** PageRank Algorithm.

**Algorithm:** PageRank

**Input:** Set of pages $P$, damping factor $\alpha$, convergence threshold $\epsilon$
**Output:** PageRank scores for each page
Initialize $PR(p) \leftarrow \frac{1}{|p|}$ for all pages $p \in P$
**repeat**
  **for** each page $p \in P$ **do**
    $PR(p) \leftarrow (1 - \alpha) / |p|$
    $PR(p) \leftarrow PR(p) + \alpha \sum_{p_j \in In(p)} \frac{PR(p_j)}{L(p_j)}$
  **end for**
  $TotalRankScore \leftarrow \sum_{P \in P} PR(p)$
  **for** each page $p \in P$ **do**
    $PR(p) \leftarrow \frac{PR(p)}{TotalTrustScore}$
  **end for**
  **Check for convergence:**
  **if** $|PR(p) - PR'(p)| < \epsilon$ for all $p \in P$ **then**
    **Stop**
  **end if**
**until** convergence
**Return** *PR*

**Table 4:** TrustRank Algorithm.

**Algorithm:** TrustRank

**Input:** Set of pages $P$, decay factor $\alpha$, set of seed (trusted) pages T, convergence threshold $\epsilon$
**Output:** TrustRank scores for each page
Define $S(p)$ as a function:

$$S(p) = \begin{cases} 1, & if\ p \in T \\ 0, & if\ p \notin T \end{cases}$$

**repeat**
  **for** each page $p \in P$ **do**
    $TR(p) \leftarrow (1 - \alpha) \cdot S(p) + \alpha \sum_{p_j \in In(p)} \frac{TR(p_j)}{L(p_j)}$
  **end for**
  $TotalTrustScore \leftarrow \sum_{p \in P} TR(p)$
  **for** each page $p \in P$ **do**
    $TR(p) \leftarrow \frac{TR(p)}{TotalTrustScore}$
  **end for**
  **Check for convergence:**
  **if** $|TR(p) - TR'(p)| < \epsilon$ for all $p \in P$ **then**
    **Stop**
  **end if**
**until** convergence
**Return** *TR*

## 5.3 Algorithmic Trustworthiness Assessment of AI Systems

Before assessing the trustworthiness of an AI system, it is essential to understand its architecture and components. The components of an AI system may vary depending on the organization that develops it, as well as the nature of the application and its intended use. Although there are shared conceptual foundations across different AI systems, it is important to note that each organization delineates the components and structures of the software it aims to develop in distinct ways. The proposed algorithmic approach assumes that the reliance relationships between the software components and the trustworthiness requirements of the relevant framework, in this case, the ALTAI, have already been defined. We propose applying Link Analysis algorithms to assess the trustworthiness of AI system components. By assessing these components, the overall trustworthiness of the AI system, concerning specific requirements, could be determined. The proposed methodological approach enables the investigation of the propagation of trustworthiness within the system and the dependency relationships between system components in terms of trustworthiness. As the conceptual foundation for our approach, we adopted the ALTAI schema, which represents a synthesis of the meta-analysis presented in Section 3, as it closely aligns with the content and findings of that analysis. Figure 3 shows the Trustworthy AI requirements and aspects of each, as described by the AI HLEG [7, 20]. These requirements and aspects serve as part of the elements of the graph-based algorithmic approach presented in what follows.

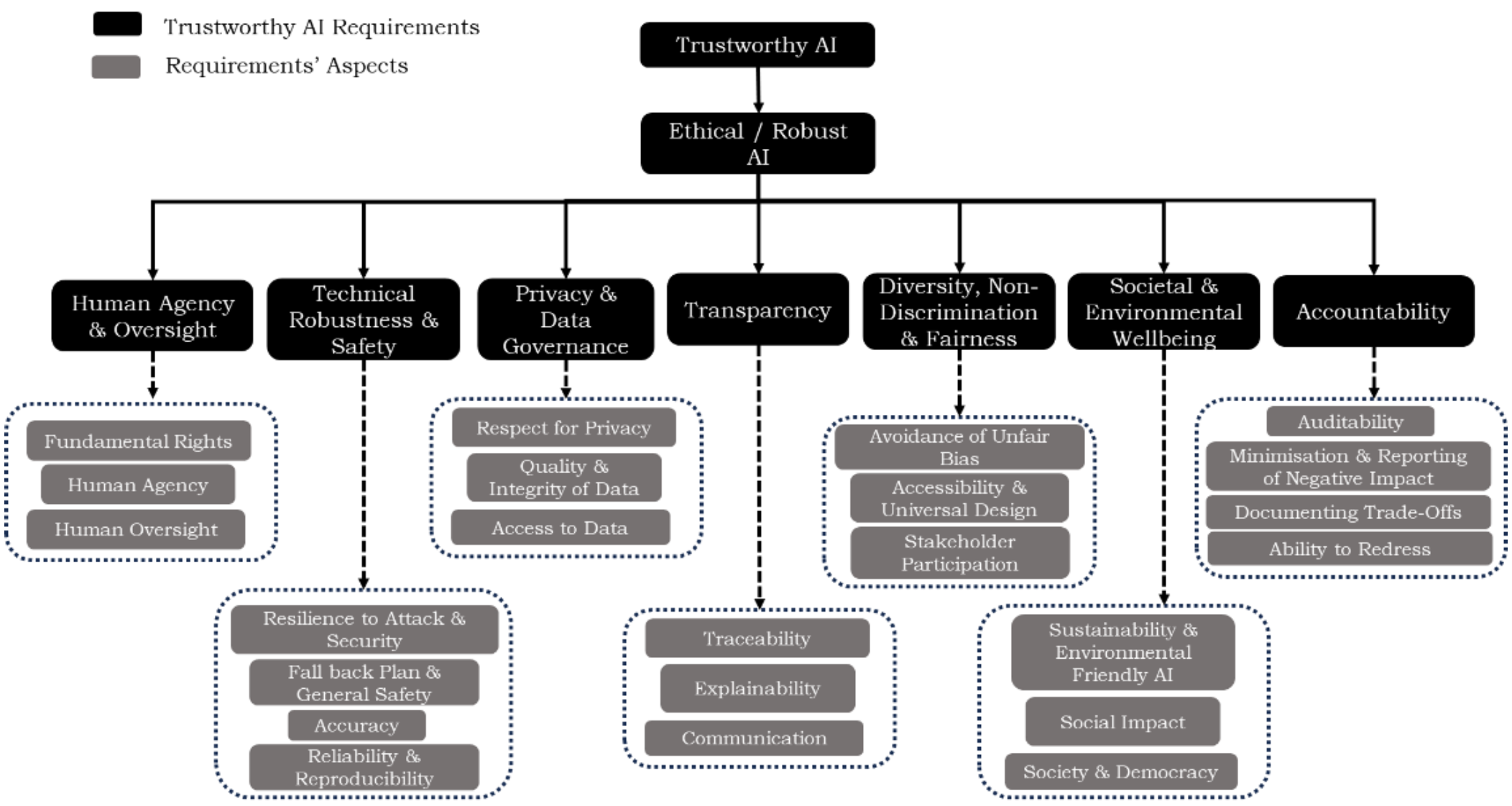

**Figure 3:** The ALTAI Requirements and Their Aspects according to the AI HLEG.

## A top-down approach

The components of an AI system can be derived from its structure and linked to data, user input, and system outputs [49, 20]. Additionally, the user interaction environment and the environments affected by the system (virtual or physical) can serve as sources for identifying various components related to the requirements for Trustworthy AI. Therefore, each aspect of every trustworthiness requirement is assumed to be associated with certain components of the AI system. Based on this logic, we construct a directed graph that links each aspect $A_j$ ( $j = 1,\ldots,n$ ) of every trustworthiness requirement $R_i$ ( $i = 1,\ldots,m$ ) to the relevant components $M_w$ ( $w = 1,\ldots,k$ ) of the AI system. We also note that there may be interconnections between aspects and components themselves. Figure 4 presents an overview of the structural organization of the graph representing our hypothetical system. It illustrates the connections within the scope of a Trustworthy AI requirement $R_i$. It is evident that the depth of the system representation can be configured to encompass multiple levels. It is proposed to adopt a two-level depth representation, comprising the following categories-entities:

**[ Requirement $R_i$ ⟶ Requirement Aspects $A_j$ ⟶ Aspects Components $M_w$ ].**

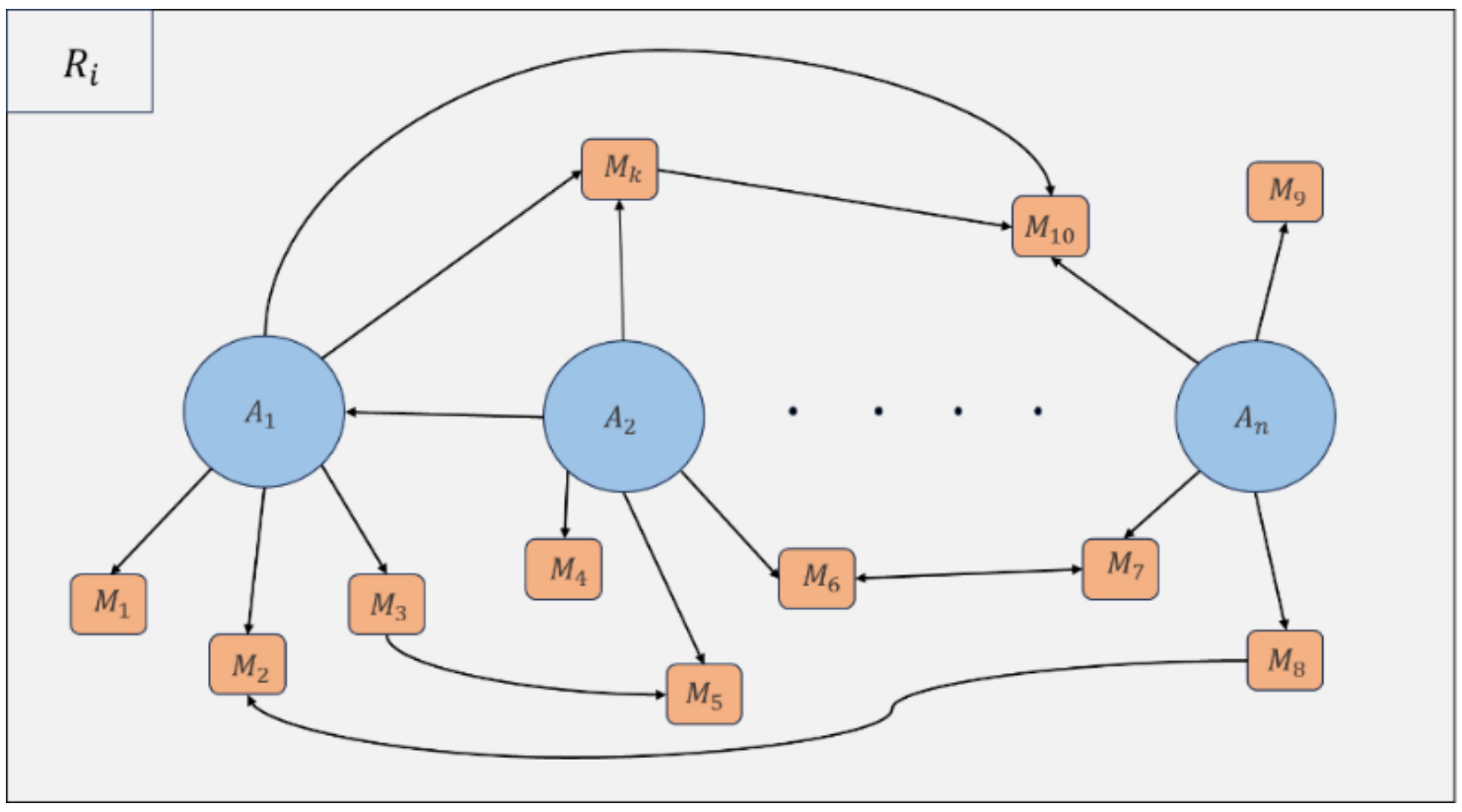

**Figure 4:** Abstract Representation of Aspects and Components and Their Connections for a Trustworthy AI Requirement (top-down approach).

The aforementioned pattern of connections can be described as a top-down approach. In other words, Requirements are linked to their respective Aspects, which are connected to the Components of the AI system they affect. This process occurs in a manner that progresses from the more general entity to the more specific one. Since the PageRank and TrustRank algorithms assign higher scores to nodes at the end of directed links, it is reasonable to assume that entities positioned lower in a top-down hierarchy are likely to receive higher scores, especially in TrustRank. Accordingly, the above-mentioned approach provides a more comprehensive understanding of the degree of trustworthiness associated with the Components of the Aspects of Requirements.

### A bottom-up approach

The diametric case implies a bottom-up approach and is expressed through a pattern of connections, which is as follows:

**[ Aspects Components $\boldsymbol{M_w}$ ⟶ Requirement Aspects $\boldsymbol{A_j}$ ⟶ Requirement $\boldsymbol{R_i}$ ]**.

Figure 5 illustrates this case at the abstract level. By applying the PageRank and TrustRank algorithms to a graph that follows the bottom-up approach (i.e., from the most specific to the most general), we can obtain scores that more accurately reflect the degree of trustworthiness of the higher-level nodes (e.g., nodes of the Aspects of Requirements). Furthermore, this approach allows us to ascertain the trustworthiness of the Requirement, as we obtain more representative scores for its various Aspects. A higher score assigned to a node indicates that it is expected to perform in the desired or predicted manner, aligning with the philosophical perspective on trustworthiness discussed earlier. The application of a top-down approach, or its opposite, provides a framework for interpreting these scores as indicators of the expected performance of the entities within the graph structure. This approach implicitly adopts the philosophical concept of trustworthiness. Essentially, the objective is to determine the trustworthiness of the AI system through algorithmic methodologies, whose outcomes also seem relevant to the previously outlined philosophical perspective.

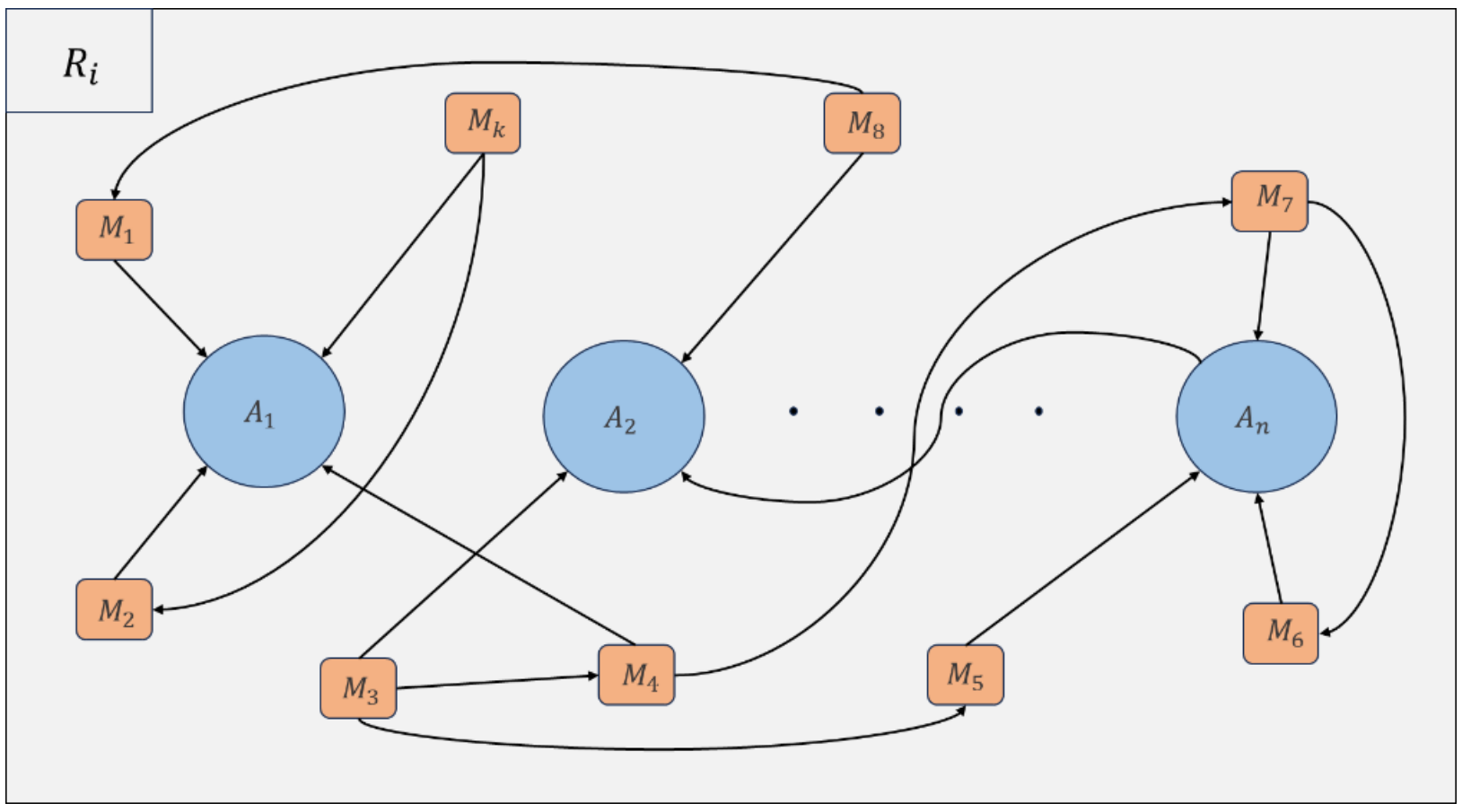

**Figure 5:** Abstract Representation of Aspects and Components and Their Connections for a Trustworthy AI Requirement (bottom-up approach).

## The philosophical view of algorithms

It is worthwhile to briefly explore the philosophical perspective on AI trustworthiness concerning algorithmic methods like PageRank and TrustRank. While these techniques originate in computational contexts, they reveal a deeper alignment with the philosophical understanding of trust as an attitude grounded in expectation, vulnerability, and moral hope. For example, if node $A_1$ links to $M_{10}$ in a graph, this implies a reliance on $M_{10}$ to behave as expected, mirroring the philosophical notion that trust involves more than dependence; it includes a normative belief that the other entity ought to act accordingly. PageRank, by attributing importance to nodes based on incoming links, can be seen as quantifying this collective anticipation. A higher PageRank score suggests that many entities in the network expect a node to perform reliably, thus elevating its status as a potentially trustworthy agent. TrustRank makes this implication even more explicit. When a node is included in the trusted seed set and assigned a trust score of 1, we are not merely acknowledging its technical connectivity. We are also expressing a belief, much like the moral commitment found in philosophical accounts, that the node should behave in accordance with the standards of trustworthiness. The propagation of trust scores throughout the network mimics the diffusion of confidence among social actors, shaped by both direct experience and structural relationships.

Importantly, both algorithms implicitly depend on a notion of trust that goes beyond calculative prediction. They incorporate the kind of 'extra factor' discussed in philosophical literature, which refers to the underlying reason why a trustor would choose to entrust a trustee with carrying out a task faithfully. As links accumulate toward a node, so does the collective expectation that the node will fulfill the relational dependencies represented by those connections. Thus, the philosophical concept of trustworthiness, understood as a property grounded in competence, reliability, and integrity, finds a compelling analogue in these algorithmic approaches. To summarize, PageRank and TrustRank are grounded in the assumption that relational structure can reveal behavioral or operational expectations. In doing so, they instantiate a formalization of trust that aligns with interdisciplinary definitions. Rather than serving as abstract parallels, these algorithms represent a computational instantiation of philosophical trust, where reliance and normative belief converge within technologically mediated interactions.

## Scenario 1: Technical Robustness and Safety (top-down approach)

To assess the trustworthiness of an AI system concerning the Technical Robustness & Safety requirement, the system can be conceptualized in the form of a graph, depicted in Figure 6. In this theoretical framework, algorithms were applied to this graph structure, generating the results summarized in Table 5. Trusted nodes within the graph were identified based on the terms enclosed within curly brackets. The outcomes of these algorithmic analyses are presented graphically in Figures 7 - 9, providing a visual representation of the propagation of trust within the system.

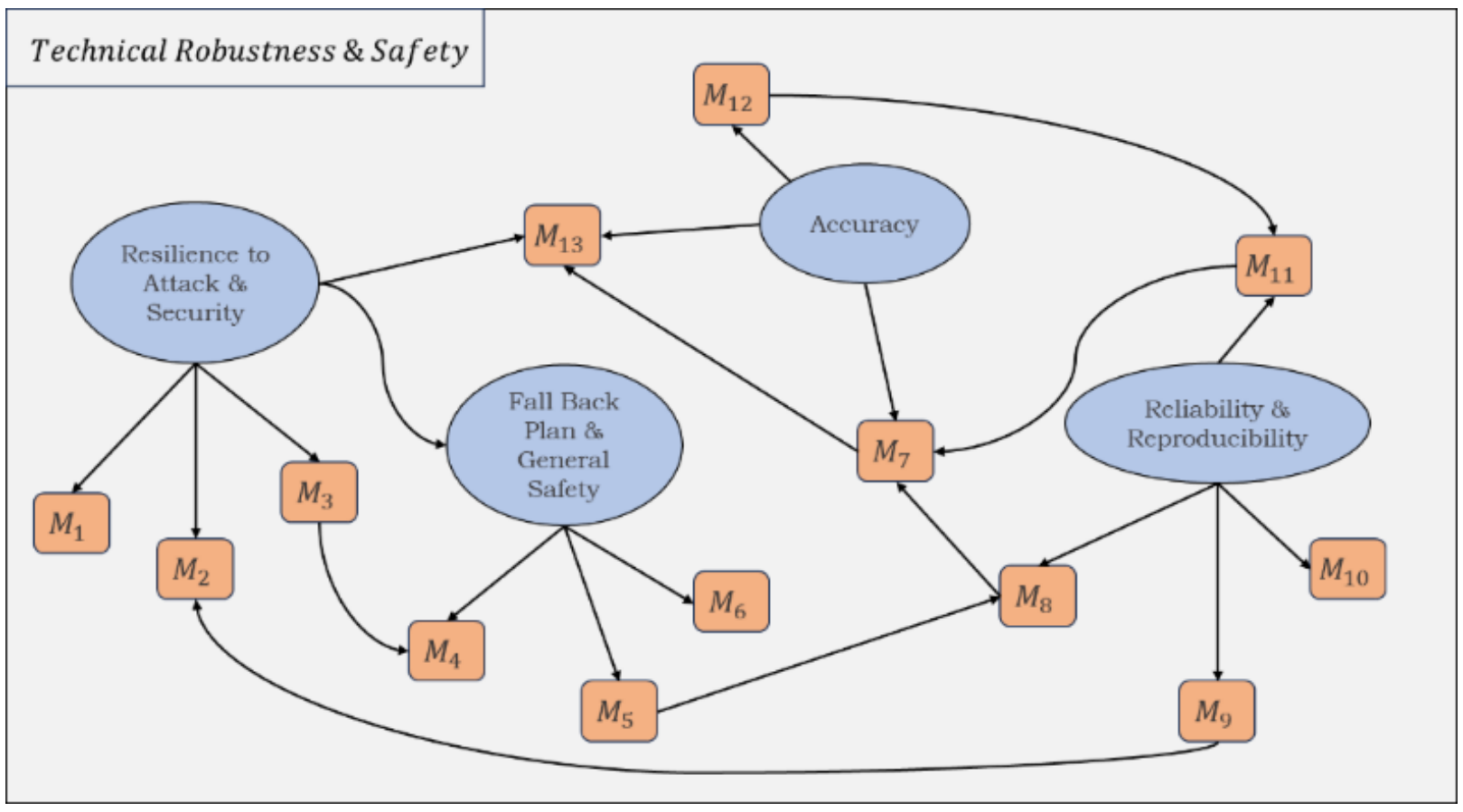

**Figure 6:** A Hypothetical Example of the Interdependencies Among Components for the Requirement 'Technical Robustness & Safety' (top-down approach).

**Table 5:** Aspects' and Components' Scores Related to Technical Robustness & Safety (about Figure 6).

| **Aspects $A_j$ and Components $M_k$** | **Scores** | | | | | | |
|---|---|---|---|---|---|---|---|
| | **PageRank** | **TrustRank** $\{A_1, A_2, A_3, A_4\}$ | **TrustRank** $\{A_3, A_4\}$ | **TrustRank** $\{A_2, A_4\}$ | **TrustRank** $\{A_1, A_3, M_5, M_{11}\}$ | **TrustRank** $\{A_1, A_2, A_3, M_4, M_6, M_7, M_9\}$ | **TrustRank** {All Nodes} |
| $A_1$ | 0.0157 | 0.0987 | 0 | 0 | 0.0933 | 0.0761 | 0.0301 |
| $A_2$ | 0.0205 | 0.1155 | 0 | 0.1934 | 0.0158 | 0.0891 | 0.0352 |
| $A_3$ | 0.0157 | 0.0987 | 0.1848 | 0 | 0.0933 | 0.0761 | 0.0301 |
| $A_4$ | 0.0157 | 0.0987 | 0.1848 | 0.1934 | 0 | 0 | 0.0301 |
| $M_1$ | 0.0205 | 0.0167 | 0 | 0 | 0.0158 | 0.0129 | 0.0352 |
| $M_2$ | 0.0534 | 0.0346 | 0.0333 | 0.0349 | 0.0158 | 0.0777 | 0.0663 |
| $M_3$ | 0.0205 | 0.0167 | 0 | 0 | 0.0158 | 0.0129 | 0.0352 |
| $M_4$ | 0.0572 | 0.0470 | 0 | 0.0548 | 0.0179 | 0.1124 | 0.0700 |
| $M_5$ | 0.0261 | 0.0327 | 0 | 0.0548 | 0.0978 | 0.0252 | 0.0401 |
| $M_6$ | 0.0261 | 0.0327 | 0 | 0.0548 | 0.0044 | 0.1014 | 0.0401 |
| $M_7$ | 0.2041 | 0.1075 | 0.1570 | 0.1095 | 0.1955 | 0.1316 | 0.1576 |
| $M_8$ | 0.0613 | 0.0488 | 0.0392 | 0.0877 | 0.0831 | 0.0214 | 0.0706 |
| $M_9$ | 0.0217 | 0.0209 | 0.0392 | 0.0411 | 0 | 0.0761 | 0.0365 |
| $M_{10}$ | 0.0217 | 0.0209 | 0.0392 | 0.0411 | 0 | 0 | 0.0365 |
| $M_{11}$ | 0.0576 | 0.0447 | 0.0838 | 0.0411 | 0.1158 | 0.0183 | 0.0693 |
| $M_{12}$ | 0.0237 | 0.0279 | 0.0523 | 0 | 0.0264 | 0.0215 | 0.0386 |
| $M_{13}$ | 0.3380 | 0.1362 | 0.1858 | 0.0930 | 0.2085 | 0.1464 | 0.1778 |

$A_1$: Resilience to Attack & Security, $A_2$: Fall Back Plan & General Safety, $A_3$: Accuracy, $A_4$: Reliability & Reproducibility

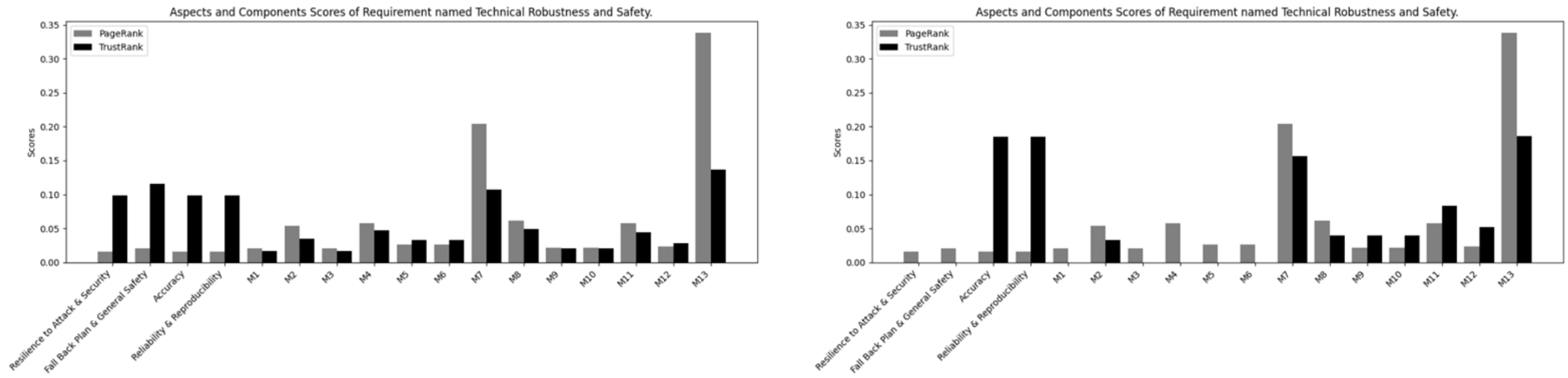

**Figure 7:** PageRank and TrustRank Results $[\{A_1, A_2, A_3, A_4\}, \{A_3, A_4\}]$.

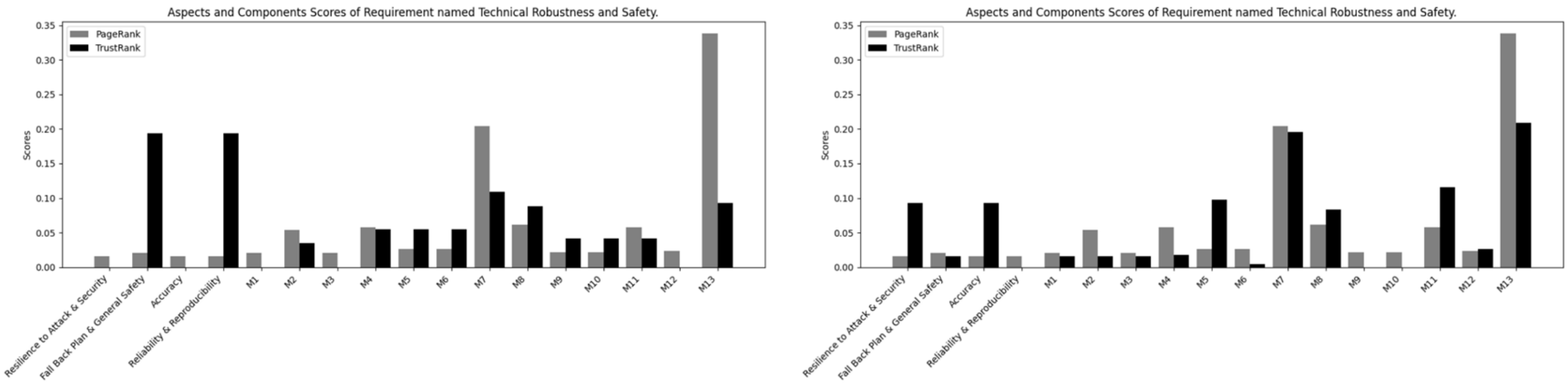

**Figure 8:** PageRank and TrustRank Results $[\{A_2, A_4\}, \{A_1, A_3, M_5, M_{11}\}]$.

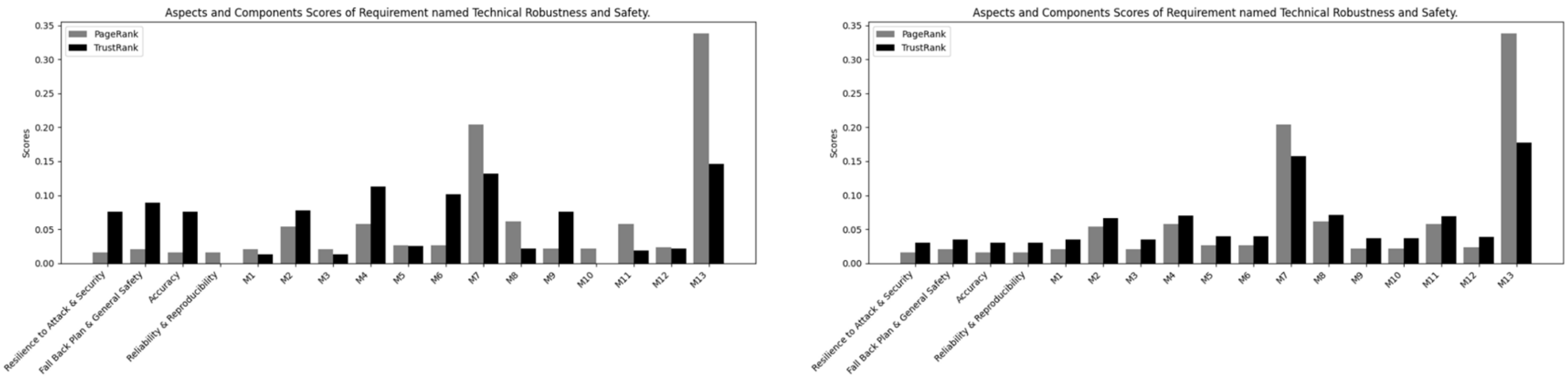

**Figure 9:** PageRank and TrustRank Results $[\{A_1, A_2, A_3, M_4, M_6, M_7, M_9\}, \{\text{All Nodes}\}]$.

### Scenario 2: Transparency (bottom-up approach)

Another paradigm is presented in Figure 10, with the corresponding results displayed in Table 6. This figure illustrates a graph constructed using the bottom-up approach. Graphically, we can see the results of the algorithms in Figures 11 - 13. This hypothetical example serves to elicit the trustworthiness of the AI system by examining how its components align with the Transparency requirement, highlighting critical aspects of Traceability, Explainability, and Communication. It is important to recognize that the connection between ethical requirements and the corresponding aspects of an AI system is shaped by the specific ethical framework adopted and the interpretation applied to it. Additionally, the various components of the system may not only influence particular aspects but also exhibit functional dependency relationships with other components. This approach emphasizes analyzing the AI system through the hierarchical relationships between entities, moving from more specific elements (e.g., components), to broader ethical aspects, thereby capturing the structure of reliance and influence within the system.

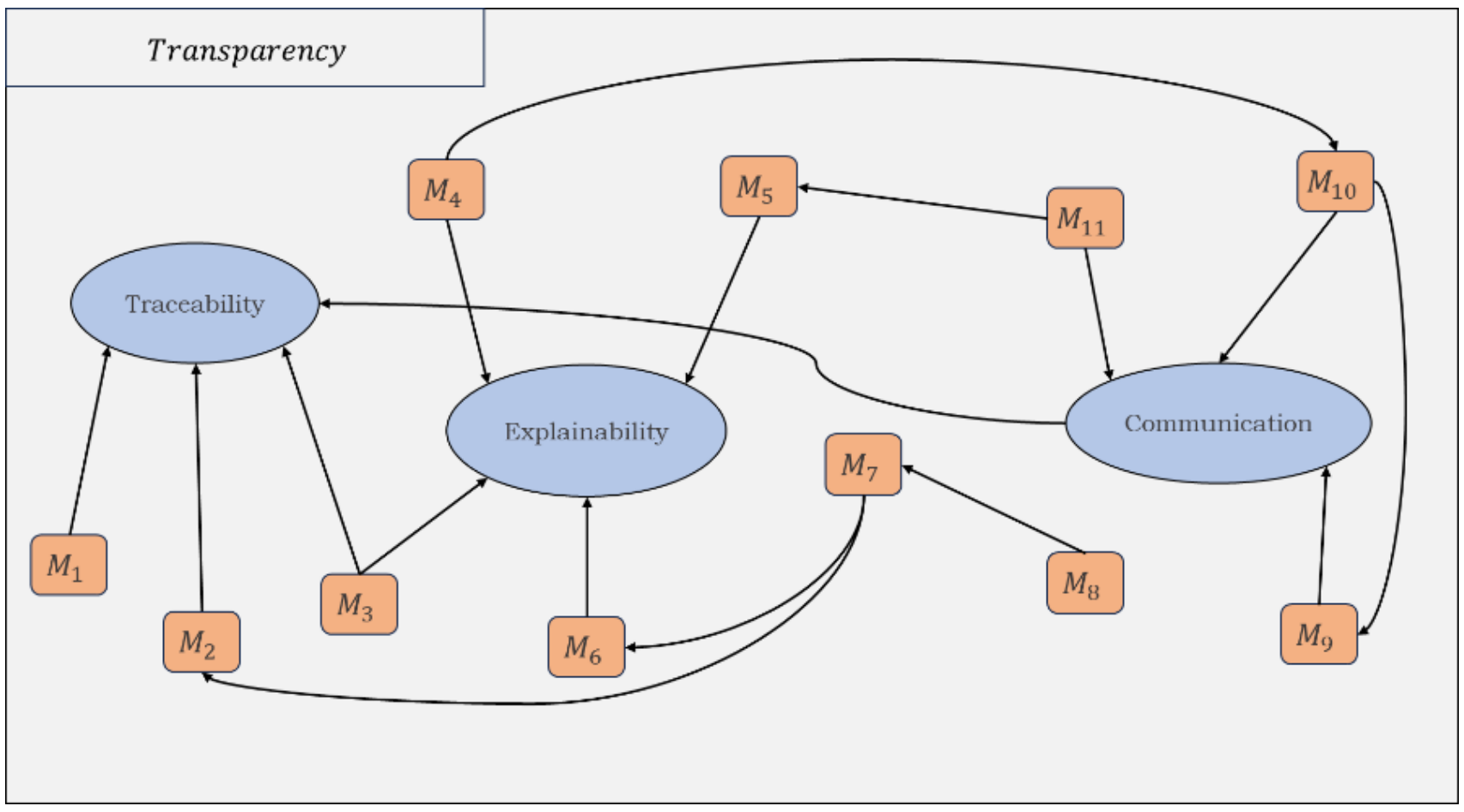

**Figure 10:** A Hypothetical Example of the Interdependencies Among Components for the Requirement 'Transparency' (bottom-up).

**Table 6:** Aspects' and Components' Scores Related to Transparency (about Figure 10).

| Aspects $A_j$ and Components $M_k$ | Scores | | | | | | |
|---|---|---|---|---|---|---|---|
| | PageRank | TrustRank $\{A_1, A_2. A_3\}$ | TrustRank $\{M_3, M_{4,} M_5\}$ | TrustRank $\{A_3, M_3, M_{4,} M_5\}$ | TrustRank $\{M_3, M_4, M_5, M_{10,} M_{11}\}$ | TrustRank $\{M_3, M_5, M_7, M_9, M_{11}\}$ | TrustRank {All Nodes} |
| $A_1$ | 0.3372 | 0.4805 | 0.1116 | 0.1901 | 0.1473 | 0.1637 | 0.2261 |
| $A_2$ | 0.1927 | 0.2597 | 0.2677 | 0.2073 | 0.1746 | 0.1749 | 0.1550 |
| $A_3$ | 0.1182 | 0 | 0.0526 | 0.1627 | 0.1309 | 0.1116 | 0.1149 |
| $M_1$ | 0.0195 | 0 | 0 | 0 | 0 | 0 | 0.0338 |
| $M_2$ | 0.0579 | 0 | 0 | 0 | 0 | 0.0372 | 0.0604 |
| $M_3$ | 0.0195 | 0 | 0.1575 | 0.1219 | 0.0847 | 0.0875 | 0.0338 |
| $M_4$ | 0.0195 | 0 | 0.1575 | 0.1219 | 0.0847 | 0 | 0.0338 |
| $M_5$ | 0.0345 | 0 | 0.1575 | 0.1219 | 0.1207 | 0.1248 | 0.0482 |
| $M_6$ | 0.0579 | 0 | 0 | 0 | 0 | 0.0372 | 0.0604 |
| $M_7$ | 0.0496 | 0 | 0 | 0 | 0 | 0.0875 | 0.0626 |
| $M_8$ | 0.0195 | 0 | 0 | 0 | 0 | 0 | 0.0338 |
| $M_9$ | 0.0195 | 0 | 0.0284 | 0.0220 | 0.0513 | 0.0875 | 0.0543 |
| $M_{10}$ | 0.0345 | 0 | 0.0669 | 0.0518 | 0.1207 | 0 | 0.0482 |
| $M_{11}$ | 0.0195 | 0 | 0 | 0 | 0.0847 | 0.0875 | 0.0338 |

$A_1$: Traceability, $A_2$: Explainability, $A_3$: Communication

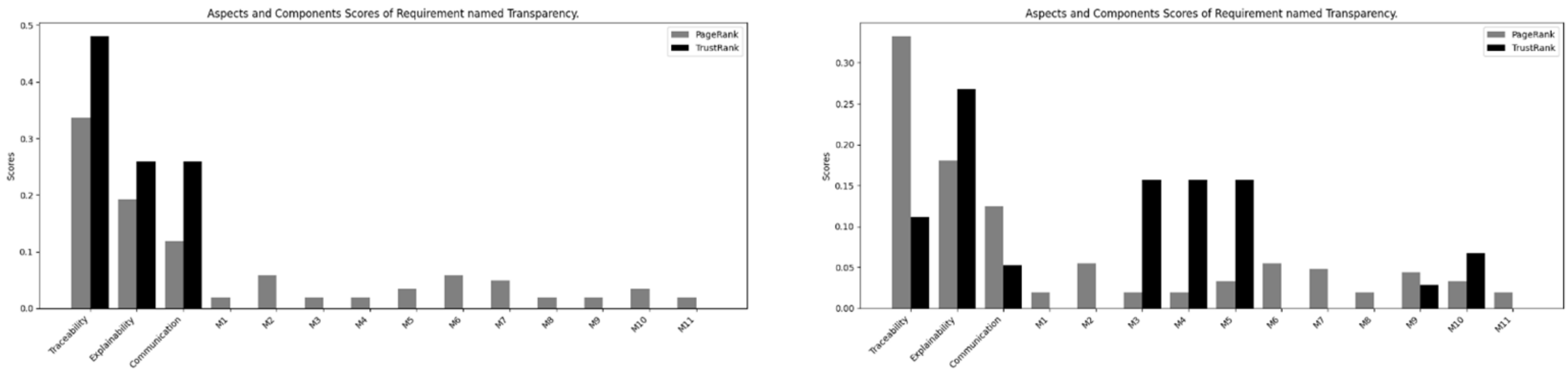

**Figure 11:** PageRank and TrustRank Results [$\{A_1, A_2, A_3\}$, $\{M_3, M_4, M_5\}$].

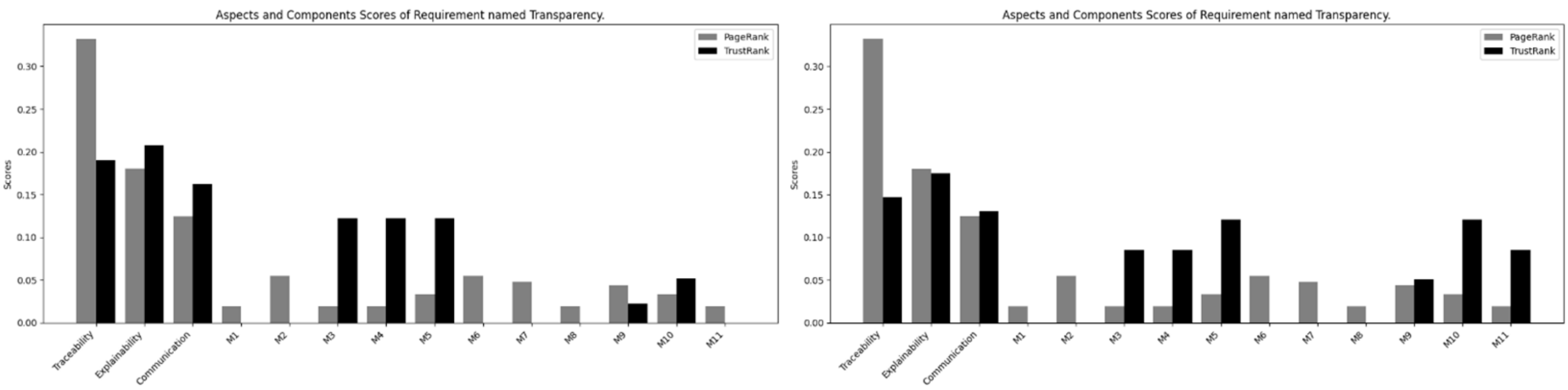

**Figure 12:** PageRank and TrustRank Results [$\{A_3, M_3, M_4, M_5\}$, $\{M_3, M_4, M_5, M_{10,} M_{11}\}$].

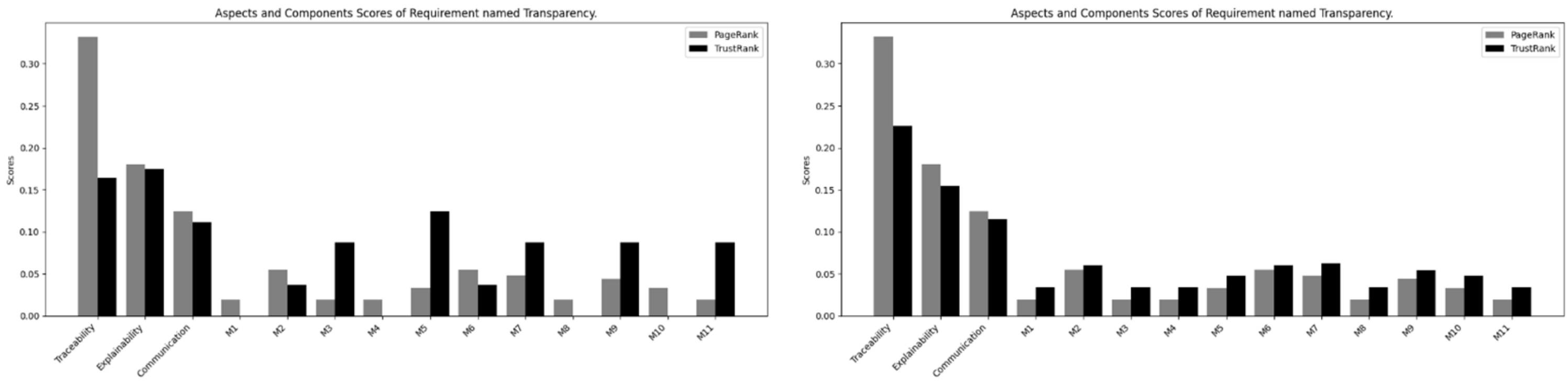

**Figure 13:** PageRank and TrustRank Results [$\{M_3, M_5, M_7, M_9, M_{11}\}$, {All Nodes}].

## Synthesis of Key Findings

In the previously discussed examples, the outcomes of the PageRank and TrustRank algorithms were obtained for the graphs presented in Figures 6 and 10. Most of the time, a clear distinction is evident between the scores obtained by the two algorithms. However, there are instances where the scores exhibit a high degree of similarity. The cases can be included in the following general conditions:

(a) $PR_{score} >> TR_{score}$,
(b) $PR_{score} << TR_{score}$,
(c) $PR_{score} \approx TR_{score}$.

If the scores of a component satisfy condition (a), then it can be classified as critical, central and not well-connected to trusted nodes. These components need to be monitored to verify the trustworthiness of their operations. In the case where a component satisfies condition (b), then it can be classified as less critical but trustworthy. It may not require immediate monitoring but it could serve as a standard for trust. Condition (c) is characterized by a dual status. The two scores can be similar, taking either low or high values. If the two scores of a component are both similar and low, this component can be regarded as less critical and weakly connected to trusted nodes. In such a case, it is appropriate to try to improve the trustworthiness of the component and to reflect on whether it can be omitted from the system or not. Conversely, in cases where the scores are both similar and high, the corresponding component may be regarded as critical and trustworthy. In the previous case, which is the optimal one among those described, it is sufficient to supervise the component and endeavor to maintain its trustworthiness level, as it seems to be of central importance.

It is essential to mention that the choice of the direction of the connections within the graph representing the AI system significantly affects the interpretation of the results. As previously noted, in the top-down approach, entities at lower levels (i.e., system components) are expected to exhibit higher scores. This enables a clearer assessment of the trustworthiness of components within the AI system, as it provides evidence of the propagation of trustworthiness from higher-level entities to their corresponding lower-level counterparts. Antithetically, the bottom-up approach exhibits a tendency for higher-level entities, such as aspects of AI trustworthiness requirements, to achieve higher scores. This allows us to gather evidence for the trustworthiness of the system concerning overarching aspects, such as Social Impact, linked to requirements like Societal and Environmental Wellbeing. In conclusion, the proposed algorithmic approach enables the determination of the AI system's trustworthiness level by analyzing the interdependencies among the entities embedded within its structure. Furthermore, it offers insights into the propagation of trust throughout the system, thereby allowing for sensitivity analysis to assess how trust shifts in response to changes in the trust status of specific entities within the AI system.

## 6. Conclusions, Limitations, and Prospects

In this paper, we demonstrated how Link Analysis algorithms, such as PageRank and TrustRank, can be utilized as complementary techniques for exploring the trustworthiness of an AI system. A key step in our approach was linking components of the system to specific trustworthiness requirements. The ALTAI framework was employed to identify these requirements, as it showed strong alignment with findings from our meta-analysis of systematic and semi-systematic literature reviews. The AI system was represented as a graph, with edges denoting the expectation of trustworthy behavior between components with respect to defined ethical criteria.

Building on this, we developed an exploratory approach to bridge theoretical definitions of trustworthiness with their algorithmic representation. More specifically, we proposed that directed connections within the graph reflect not only functional dependencies but also a normative anticipation, one that aligns with the philosophical notion of trust as a reliance shaped by competence, reliability, and integrity. Thus, our framework integrates both the computational and the conceptual dimensions: PageRank and TrustRank are not just mathematical tools but could be instantiations of a deeper model of trust, one that encompasses both operational and ethical expectations. Our results indicate that these algorithms can inform us about two central aspects of AI trustworthiness. First, they reveal how certain components depend on others to fulfill expected behavior. Second, they show how trustworthiness propagates through the system, helping to identify the most trustworthy components, those critical to system functionality, and those less central. This algorithmic approach offers a less subjective method for assessing trustworthiness, helping to reduce reliance on purely qualitative or self-assessment methods.

Nevertheless, several limitations remain. Empirically, the lack of detailed data on the architecture of real-world AI systems constrains the accuracy of our graph models. Theoretically, our analysis remains static and does not account for dynamic behavior over time. Additionally, structural connectivity alone cannot fully capture the nuanced and multidimensional character of trustworthiness, which often involves context-sensitive, ethical, and human-centered factors. Future work should focus on incorporating dynamic system behaviors, more detailed architectural data, and insights from socio-technical and interdisciplinary approaches to AI evaluation. Additionally, there is a critical need for a comprehensive conceptual model or a well-documented catalogue that captures use cases and cross-references these insights, providing a structured foundation for understanding and aligning system design with real-world contexts. Through this exploratory approach, we can move toward a more holistic and rigorous assessment of AI trustworthiness that is both technically grounded and philosophically informed.